\documentclass[sigconf]{acmart}
\setcopyright{none}
\renewcommand\footnotetextcopyrightpermission[1]{} 
\usepackage{tikz}
\usepackage{float}
\usepackage{listings}
\usepackage{multirow}
\usepackage{hyperref}

\usepackage{comment}

\usetikzlibrary{automata, positioning, arrows}

\usepackage{fancyhdr}
\usepackage{atbegshi,picture}

\begin{document}

\title[Towards Embodied Cognition in Robots via Spatially Grounded Synthetic Worlds]{Towards Embodied Cognition in Robots via Spatially Grounded Synthetic Worlds}


\author{Joel Currie \textsuperscript{* \dag}}
\email{joel.currie@iit.it}
\orcid{0000-0002-3367-7056}
\affiliation{
  \institution{}
  \streetaddress{}
  \city{}
  \country{}
  \postcode{}
}

\author{Gioele Migno \textsuperscript{*}}
\email{gioele.migno@iit.it}
\orcid{0009-0004-0471-3064}
\affiliation{
  \institution{}
  \streetaddress{}
  \city{}
  \country{}
  \postcode{}
}

\author{Enrico Piacenti \textsuperscript{*}}
\email{enrico.piacenti@iit.it}
\orcid{0009-0000-2197-5278}
\affiliation{
  \institution{}
  \streetaddress{}
  \city{}
  \country{}
  \postcode{}
}

\author{Maria Elena \textsuperscript{\dag} Giannaccini }
\email{elena.giannaccini@abdn.ac.uk}
\orcid{0000-0002-0871-4804}
\affiliation{
  \institution{}
  \streetaddress{}
  \city{}
  \country{}
  \postcode{}
}

\author{Patric Bach \textsuperscript{\dag}}
\email{patric.bach@abdn.ac.uk}
\orcid{0000-0003-4493-2080}
\affiliation{
  \institution{}
  \streetaddress{}
  \city{}
  \country{}
  \postcode{}
}

\author{Davide De Tommaso \textsuperscript{*}}
\email{davide.detommaso@iit.it}
\orcid{0000-0003-2132-5261}
\affiliation{
  \institution{}
  \streetaddress{}
  \city{}
  \country{}
  \postcode{}
}

\author{Agnieszka Wykowska \textsuperscript{*}}
\email{agnieszka.wykowska@iit.it}
\orcid{0000-0003-3323-7357}
 \affiliation{
  \institution{}
  \streetaddress{}
  \city{}
  \country{}
  \postcode{}
}


\renewcommand{\shortauthors}{Joel Currie, Gioele Migno, Enrico Piacenti, Maria Elena Giannaccini, Patric Bach, Davide De Tommaso, \& Agnieszka Wykowska}
\begin{teaserfigure}
  \centering
  \includegraphics[width=1\linewidth]{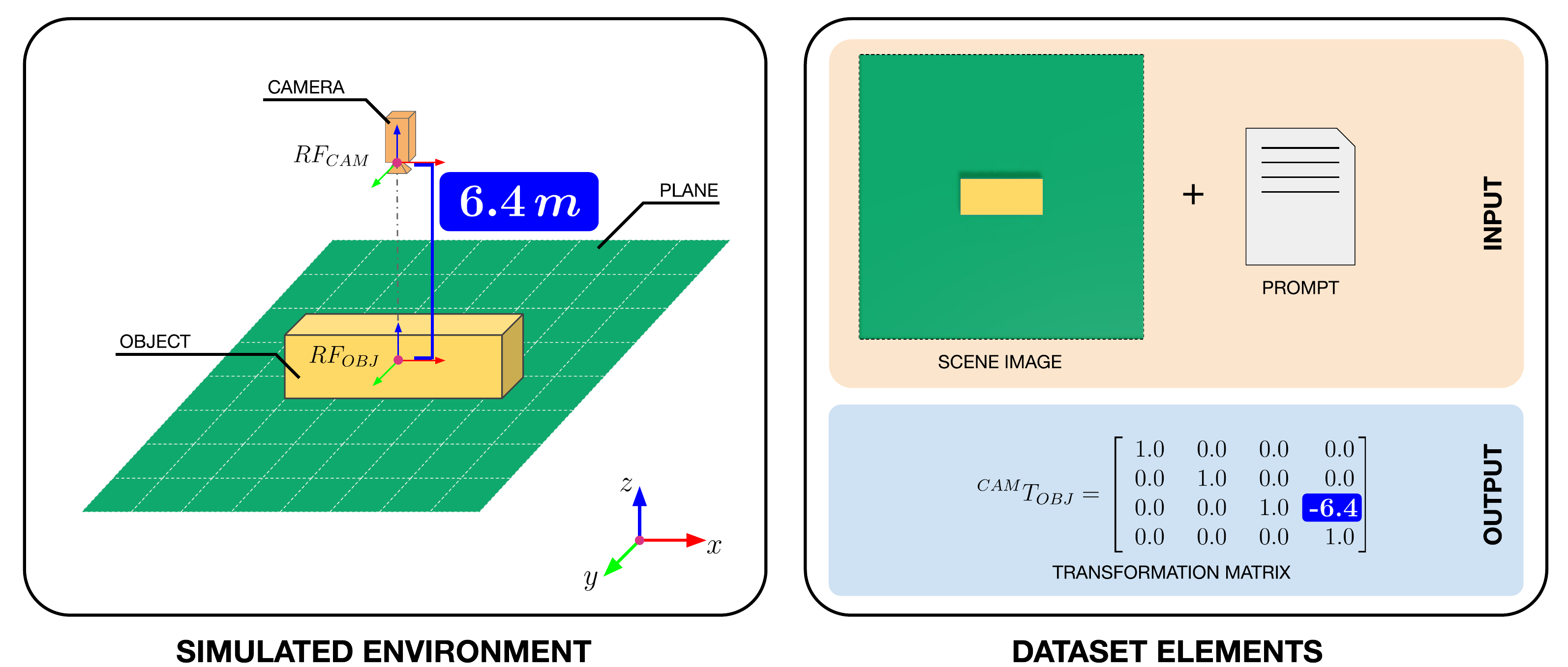}
  \caption{Synthetic environment and dataset elements. A minimal 3D scene is procedurally generated with a non-uniform scaled cube and overhead camera. Each instance yields an RGB image, a language prompt, and a 4×4 transformation matrix $\left({}^{CAM}T_{OBJ}\right)$ representing object reference frame pose $\left(RF_{OBJ}\right)$ with respect to the camera reference frame $\left(RF_{CAM}\right)$, enabling structured spatial representations for supervised learning in embodied AI.}
  \label{fig:dataset_example}
\end{teaserfigure}

\begin{abstract}
We present a conceptual framework for training Vision-Language Models (VLMs) to perform Visual Perspective Taking (VPT), a core capability for embodied cognition essential for Human-Robot Interaction (HRI). As a first step toward this goal, we introduce a synthetic dataset, generated in NVIDIA Omniverse, that enables supervised learning for spatial reasoning tasks. Each instance includes an RGB image, a natural language description, and a ground-truth 4×4 transformation matrix representing object pose. We focus on inferring Z-axis distance as a foundational skill, with future extensions targeting full 6 Degrees Of Freedom (DOFs) reasoning. The dataset is publicly available to support further research. This work serves as a foundational step toward embodied AI systems capable of spatial understanding in interactive human-robot scenarios.
\end{abstract}




\maketitle

\begingroup
\renewcommand\thefootnote{*}
\footnotetext{Social Cognition in Human-Robot Interaction Unit, Italian Institute of Technology, Genova, Italy}
\addtocounter{footnote}{1}
\renewcommand\thefootnote{\dag}
\footnotetext{University of Aberdeen, Aberdeen, United Kingdom}
\endgroup

\pagestyle{plain}

\section{Introduction}
Effective Human-Robot Interaction (HRI), like human-human interaction requires a suite of socio-cognitive capacities \cite{lemaignan_artificial_2017, natarajan2023human}. Among these, Visual Perspective Taking (VPT) - the capacity to infer what another sees from their point of view - plays a critical role \cite{currie_more_2024, currie_mind_2024, DOGAN2020}. VPT is foundational to many downstream interaction capabilities, including joint action \cite{Freundlieb2016}, social navigation \cite{Kozhevnikov2006} and mental/affective/goal state inference \cite{Batson1997, Mattan2016, Furlanetto2016}. Consider a toy example: you ask a collaborator \textit{"Can you pass me the object to the left?"}. To achieve the desired action the collaborator must not only identify the referenced object but also reason about the spatial relationships from distinct viewpoints, their own and yours. This requires the ability to represent how the world appears from another agent's perspective, and to map effectively between diverging frames of reference. 

Existing VPT solutions in robotics often rely on explicit geometric modelling \cite{marin2008geometric, fischer_markerless_2016} and hand-crafted perspective transformations - typically through rule-based \cite{trafton_enabling_2005} or spatial reasoning pipelines \cite{DOGAN2020}. While these methods are effective in constrained environments, they often lack flexibility, generalisability and scalability necessary for real-world HRI. By contrast, Vision Language Models (VLMs) are a method demonstrating impressive flexibility \cite{chen2024spatialvlm}, performing well in tasks such as scene understanding \cite{goral_seeing_2024}.

However, despite these strengths, current VLM's struggle with precise spatial reasoning, especially when inferring precise object poses, relative orientations or viewpoint-specific relations \cite{song2025robospatial, goral_seeing_2024, gao_vision_2025}. Recent findings have suggested this deficit in spatial reasoning is not a limitation in model architecture, but instead likely to be due to lack of training data that explicitly ties spatial relationships to grounded, visual scenes \cite{song2025robospatial, chen2024spatialvlm, Luo2025}. 
Simulated environments offer a promising solution for generating scalable datasets trivially, as large datasets are often a bottleneck in VLM training. More importantly, they also act as a proxy for embodiment, allowing the reduction of error between inferred representations and reality by enabling supervised learning from generated synthetic data in which structured spatial relationships are easily extractable and inherently exactly precise.

We contribute an early-stage framework for training VLMs to perform embodied cognitive tasks such as VPT, grounded in spatial reasoning. As a first step toward this vision, we present a proof-of-concept dataset \cite{joel_currie_2025} composed of simple synthetic scenes with ground-truth transformation matrices. Our approach aims to support the future development of spatially aware, embodied robots capable of understanding 
`what/how others see' and 
`where an object is relative to me/others'.

\section{Method}
We propose a conceptual pipeline for training VLMs to perform VPT and other embodied spatial reasoning tasks in HRI. The overarching goal is to develop a system that, given a single RGB image and a natural-language prompt describing an object, can infer its full 6 Degrees of Freedom (DOFs) pose relative to both the frame of the robot's viewpoint and that of another agent in the environment. As an initial step, we present a proof-of-concept synthetic dataset \cite{joel_currie_2025}, procedurally generated using NVIDIA Omniverse Replicator \cite{ahmed2024systemic}, containing simple 3D scenes (see Figure \ref{fig:dataset_example}). Each scene includes a single cube with randomised dimensions and material properties, a static object position, and a virtual camera with randomised height (Z-axis translation). Ground-truth transformation matrices provide precise supervision for object-to-camera pose. 

The current dataset \cite{joel_currie_2025} 
 targets a simplified version of the full task: inferring object translation along the Z-axis only, while holding rotation fixed on all axes, and X/Y translation constant. This design isolates a key spatial relation and allows for controlled evaluation of VLMs’ ability to map visual and linguistic input to structured spatial representations. Our conceptual pipeline consists of three stages: (i) object pose estimation from image-text input, yielding a transformation matrix $\left({}^{CAM}T_{OBJ}\right)$, (ii) inference of relative viewpoint transformation between an agent and the camera $\left({}^{CAM}T_{AGT}\right)$, and (iii) perspective mapping via transformation composition, producing $\left({}^{AGT}T_{OBJ}\right)$, the object’s pose from the agent’s perspective. By structuring spatial supervision in this way, we aim to advance the development of robots capable of performing embodied cognitive tasks — such as perspective taking, spatial reasoning, and viewpoint-invariant object understanding—in real-world HRI. This work lays the foundation for agents that not only perceive and describe the world but also reason about it from multiple embodied perspectives. Future work will expand the dataset to include additional DOFs, more complex scenes, and integration with robotic platforms to support real-time, perspective-aware behaviour.




\section*{Dataset Availability}
We release our synthetic dataset of minimal 3D scenes, each containing an RGB image, a natural language prompt, and a ground-truth 4×4 pose matrix. The dataset \cite{joel_currie_2025} is available at:\newline \href{https://huggingface.co/datasets/jwgcurrie/synthetic-distance}{https://huggingface.co/datasets/jwgcurrie/synthetic-distance}.

\section*{Acknowledgement}
This work has received support from the Project \textit{"Future Artificial Intelligence Research (hereafter FAIR)"}, PE000013 funded by the European Union - NextGenerationEU PNRR MUR - M4C2 - Investimento 1.3 - Avviso Creazione di \textit{"Partenariati estesi alle università, ai centri di ricerca, alle aziende per il finanziamento di progetti di ricerca di base"} CUP J53C22003010006.

\bibliographystyle{ACM-Reference-Format}

\end{document}